# Spacecraft depth completion based on the gray image and the sparse depth map

**Xiang Liu**, **Hongyuan Wang***, **Zhiqiang Yan**, **Yu Chen**

Harbin Institute of Technology, Harbin, China

**Xinlong Chen**, **Weichun Chen** [1]

China Academy of Space Technology, Beijing, China

Abstract ― Perceiving the three-dimensional (3D) structure of the spacecraft is a prerequisite for successfully executing many on-orbit space missions, and it can provide critical input for many downstream vision algorithms. In this paper, we propose to sense the 3D structure of spacecraft using light detection and ranging sensor (LIDAR) and a monocular camera. To this end, Spacecraft Depth Completion Network (SDCNet) is proposed to recover the dense depth map based on gray image and sparse depth map. Specifically, SDCNet decomposes the object-level spacecraft depth completion task into foreground segmentation subtask and foreground depth completion subtask, which segments the spacecraft region first and then performs depth completion on the segmented foreground area. In this way, the background interference to foreground spacecraft depth completion is effectively avoided. Moreover, an attention-based feature fusion module is also proposed to aggregate the complementary information between different inputs, which deduces the correlation between different features along the channel and the spatial dimension sequentially. Besides, four metrics are also proposed to evaluate object-level depth completion performance, which can more intuitively reflect the quality of spacecraft depth completion results. Finally, a large-scale satellite depth completion dataset is constructed for training and testing spacecraft depth completion algorithms. Empirical experiments on the dataset demonstrate the effectiveness of the proposed SDCNet, which achieves 0.25m mean absolute error of interest and 0.759m mean absolute truncation error, surpassing state-of-the-art methods by a large margin. The spacecraft pose estimation experiment is also conducted based on the depth completion results, and the experimental results indicate that the predicted dense depth map could meet the needs of downstream vision tasks.

**Index Terms**―Spacecraft depth completion, 3D structure recovery, Multi-source feature fusion, Satellite dataset, Deep learning

# I. INTRODUCTION

With the rapid development of aerospace technology, numerous on-orbit missions oriented to non-cooperative spacecraft have emerged. Among them, perceiving the three-dimensional structure of spacecraft and providing it to downstream vision algorithms (such as pose estimation [1]-[2], component detection [3], 3D reconstruction [4], etc.) are vital to ensure the successful execution of these tasks.

At present, stereo vision systems [5] and active time-of-flight (TOF) cameras [6] are the primary options for perceiving the three-dimension structure of non-cooperative spacecraft. Unfortunately, limited by the installation baseline length and power consumption, both of them work at close distances (generally less than 20m), which brings great challenges to space on-orbit tasks. Moreover, stereo vision systems generally work poorly on objects with smooth surfaces or repetitive textures due to their reliance on the quality of extracted feature points. Inspired by autonomous driving technology, this paper attempts to sense the three-dimensional structure of spacecraft at a long distance (maximum to 250m) using light detection and ranging sensor (LIDAR) and monocular camera. To this end, we propose a depth completion algorithm to recover the three-dimensional structure of spacecraft using a gray image and sparse depth map.

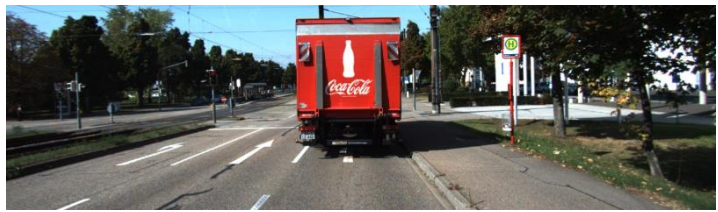

(a)

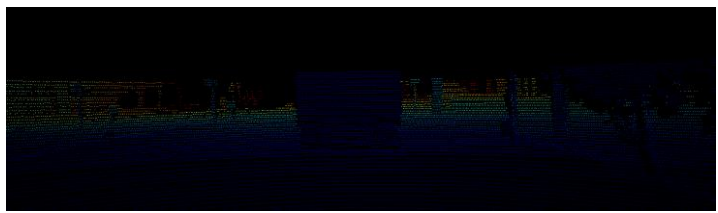

(b)

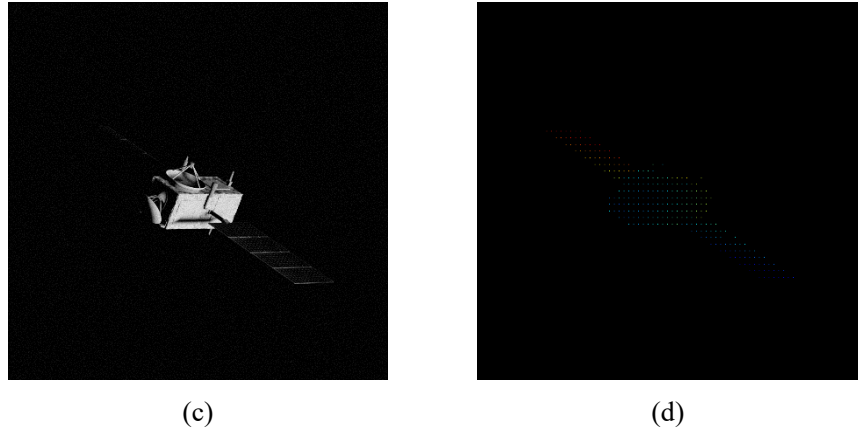

(c)                                        (d)

Fig. 1 Example Data for depth completion in different scenarios. The LIDAR ranging results are projected to image space to generate the sparse depth map. (a) the RGB image in KITTI dataset, (b) the sparse depth map in KITTI dataset, (c) the gray image of spacecraft, (d) the sparse depth map of spacecraft.

The depth completion task has been attracting considerable research due to its importance in various fields, and numerous depth completion methods for ground scenes have been proposed. Nevertheless, due to the distinct working scenarios and data characteristics, the current depth completion methods oriented to ground scenarios can't be directly applied to spacecraft depth completion. Fig.1 shows the difference between autonomous driving scene data and space on-orbit mission data. The differences in data collected from different scenarios can be embodied in the following aspects: 1) The data from ground scenarios contains rich background information. On the contrary, the on-orbit data mainly comprises the spacecraft and the simple celestial background. To some extent, the spacecraft depth completion task can be regarded as object-level depth completion. 2) Due to the different working distances, the depth map of spacecraft obtained by LIDAR is more sparse than the depth map of ground scenarios, bringing significant challenges to spacecraft depth completion. 3) Compared with the ground scene, the on-orbit lighting conditions are more complex, and some areas of the spacecraft are invisible due to lighting shadows.

To alleviate the problem, we propose a **S**pacecraft **D**epth **C**ompletion Network (**SDCNet**) for the three-dimensional structure recovery of spacecraft using a gray image and sparse depth map. Specifically, we decompose the object-level depth completion task for spacecraft into two subtasks: foreground segmentation subtask and foreground depth completion subtask, avoiding the interference of the starry background to the foreground satellite depth completion. Moreover, a multi-source feature fusion module is proposed to integrate the geometric features and context the gray image provides into the depth map feature, providing essential information guidance for spacecraft depth completion. The main contributions of this paper are as follows:

(1) A novel spacecraft depth completion framework is proposed to recover spacecraft dense depth, which decomposes the object-level depth completion into foreground segmentation and foreground depth completion, avoiding the interference of the starry background to the satellite depth completion.

(2) An attention-based feature fusion module is proposed for feature aggregation from different sources, which infers the cross-channel and spatial attention maps between features and integrates the gray image's geometric features and context information into the depth map feature.

(3) Four metrics are proposed to evaluate the quality of object-level depth completion results. MAEI and RMSEI calculate the depth error in the preserved foreground region, which more intuitively reflects the depth prediction accuracy of the object itself. MATE and RMSTE use truncation error to evaluate the depth error in the predicted foreground region, comprehensively evaluating the quality of depth completion results.

(4) We construct a satellite depth completion dataset based on 126 CAD models for training and testing spacecraft depth completion algorithms, which can promote the development of the spacecraft depth completion field.

The rest of the paper is organized as follows. Section II provides an overview of various depth completion methods. Section III elaborates on the theories and methods of the proposed spacecraft depth completion algorithm. Then, the construction method of the satellite depth completion is introduced in detail in Section IV. Experimental results in different settings and a comparison with other methods are presented in Section V. Finally, we conclude the article in Section VI.

## II. RELATED WORKS

Over the last decades, numerous depth completion algorithms with the guidance of optical images have been proposed due to the importance of depth completion in various applications, which can be roughly divided into two categories: the early fusion model and the late fusion model.

The early fusion model generally takes the concatenation of RGB-D as input and predicts dense depth through a U-Net-alike structure. For instance, Sparse-to-dense [7] concatenates the RGB images and sparse depth and adopts the encoder-decoder structure to regress depth for each pixel. Several methods [8]-[10] also extract low-level features from the optical image and depth map separately and concatenate them at the first layer of the encoder-decoder network. Moreover, the coarse-to-fine strategy is frequently adopted to generate more accurate depth estimation results [11]-[13]. S2DNet [11] utilizes the Sparse-to-dense [7] to predict the coarse-level depth map first. The estimated coarse depth map is then concatenated with the input image and fed into the fine network for fine-level depth map estimation. Many methods [13]-[18] also adopt the spatial diffusion process to refine the coarse depth map and achieve promising results. These methods utilize the encoder-decoder structure to simultaneously predict a blur depth map and the affinity matrix. Then the spatial diffusion process is used to generate a final refined depth map according to the predicted affinity matrix.

The late fusion model usually utilizes two branches to extract features from the optical image and depth map, respectively, and then perform feature fusion at intermediate layers. Inspired by guided image filtering [19], GuideNet [20] introduces the guided convolution module, which generates data-driven spatially-variant kernels from the RGB image features to integrate the edge information in RGB into depth map features. FCFRNet [21] utilizes the channel shuffle and the energy-based fusion operation to mix two features extracted from different inputs. On the basis of GuideNet [20], RigNet [22] adopts the repetitive hourglass network to generate more clear image guidance. Moreover, the repetitive guidance module is also proposed to progressively generate structure-detailed depth map features. FuseNet [23] replaces the 2D convolution in the depth map branch with 3D continuous convolution to extract 3D features of the depth map. The 3D features are then projected into image space to form a sparse depth feature map. The sparse depth feature map is finally fused with the RGB feature map for the dense depth prediction.

Although numerous depth completion algorithms oriented to the ground scene have been proposed, these methods perform poorly in the spacecraft depth completion task due to the particularity of the space on-orbit environment. Therefore, according to the characteristics of on-orbit data, a novel Spacecraft Depth Completion Network (SDCNet) is proposed for spacecraft depth completion. Specifically, SDCNet decomposes the spacecraft depth completion into the foreground segmentation subnet and foreground depth completion subnet, avoiding the interference of the starry background to the foreground spacecraft depth completion. Moreover, a novel attention-based feature fusion module is introduced to integrate the geometric features and context in the gray image into the depth map features, providing essential information guidance for spacecraft depth completion. Finally, four new metrics are proposed to evaluate the quality of object-level spacecraft depth completion results.

## III. SPACECRAFT DEPTH COMPLETION METHOD

The overall network architecture of our proposed Spacecraft Depth Completion Network (SDCNet) is illustrated in Fig. 2. The SDCNet comprises a foreground segmentation subnet (FSNet) and foreground depth completion subnet (FDCNet). The FSNet predicts the probability that each pixel belongs to the spacecraft and segments the foreground area for subsequent spacecraft depth completion. Then the FDCNet regresses the depth of the segmented foreground area by fusing gray image and sparse depth image information.

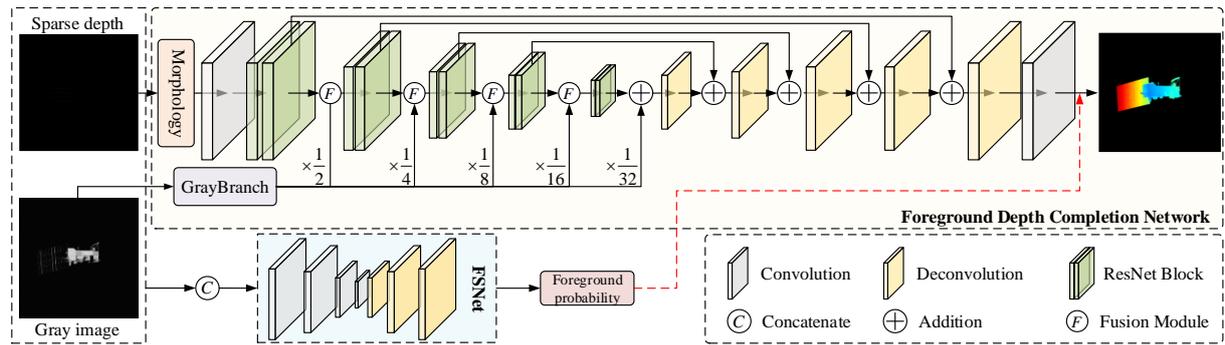

Fig. 2 The overall architecture of Spacecraft Depth Completion Network (SDCNet), which comprises a foreground segmentation subnet (FSNet) and foreground depth completion subnet (FDCNet).

A. Foreground Segmentation Subnet

Considering the on-orbit data are generally composed of spacecraft and simple celestial background, the spacecraft depth completion task can be equivalent to the object-level depth completion task. At this time, if regressing the depth of all pixels directly, the imbalance of the number of foreground and background samples will inevitably degrade network performance. Moreover, the network will also tend to learn the weight minimizing the overall pixel depth error (including the foreground and background) instead of minimizing the spacecraft depth error. To avoid these problems mentioned above, we design a simple foreground segmentation network (FSNet) to segment the foreground region, which is performed on depth regression subsequently. The detailed structure parameter of the FSNet is shown in Fig. 3.

Since the foreground segmentation subtask belongs to the pixel's binary classification problem, FSNet adopts the encoder-decoder network structure to predict the probability that each pixel belongs to the foreground. Moreover, to fully utilize the complementary information of the gray image and sparse depth map, FSNet takes their concatenation as input. Finally, the designed FSNet only contains 8 convolutional layers and 0.005M learnable parameters, which can filter out the interference of background on the subsequent spacecraft depth regression while keeping the overhead small.

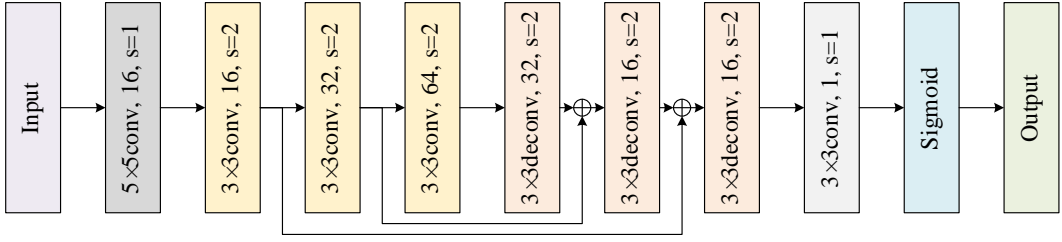

Fig. 3 The detailed structure parameter of the FSNet.

B. Foreground Depth Completion Subnet

After segmenting the foreground region, the foreground depth completion subnet (FDCNet) is designed to regress the pixel depth in the segmented foreground region. As shown in Fig. 2, FDCNet is composed of the gray image feature extraction branch and the sparse depth map completion branch, both of which adopt the encoder-decoder structure. The gray image feature extraction branch aims to extract the geometric structure and context information from the gray image, providing critical cues for the subsequent spacecraft depth prediction. At the same time, the sparse depth map completion branch predicts the pixel depth utilizing the features extracted from the gray image and the depth map. Considering the adverse effect of sparse data on convolution operations [24], we adopt simple morphological operations to preprocess the sparse depth map, generating coarse pseudo-dense depth images and feeding the result into the sparse depth map completion branch.

In the process of predict the dense depth, how effectively integrating the geometric structure and context information in the gray image into the depth map features plays a crucial role in the depth completion task. To this end, we propose a novel attention-based feature fusion module for aggregating the features extracted from the gray image and the depth map. Fig. 4 shows the specific structure of the feature fusion module, which can be divided into two stages: the cross-channel fusion stage and the spatial fusion stage. Specifically, the cross-channel fusion stage deduces the correlation between different features along the channel dimension. In contrast, the spatial fusion stage infers which feature is more informative along the spatial dimension.

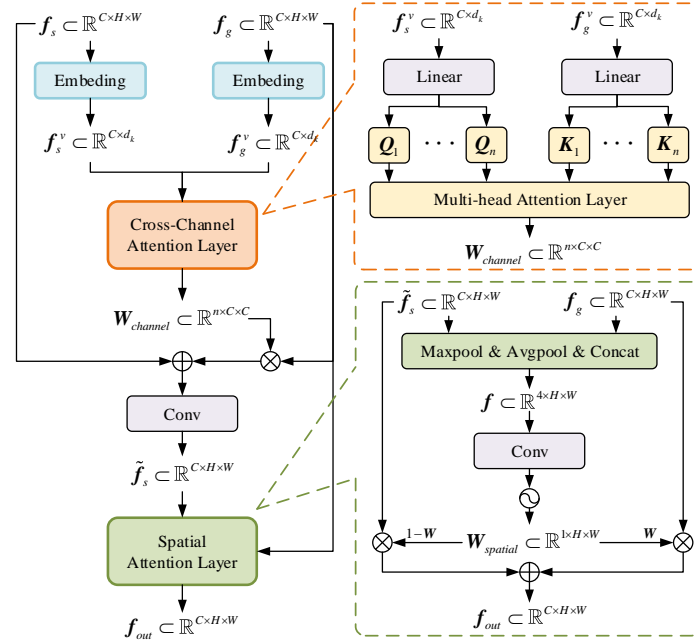

Fig. 4 The detailed structure of the feature fusion module, which is mainly composed of Cross-Channel Attention Layer and Spatial Attention Layer. The cross-channel attention layer deduces the correlation between different features along the channel dimension, while the spatial fusion layer infers which feature is more informative along the spatial dimension.

In the cross-channel fusion stage, the first step is to embed the feature map along each channel into a feature vector. Intuitively, we can directly expand the feature map along the row or the column. However, the dimension of the feature vector generated in this way is high, inevitably resulting in a large computational overhead. To avoid this problem, inspired by the Swin Transformer [25], we decompose the feature map along each channel into $M$ non-overlapping regions with size $S \times S$. The region's features under different channels are extracted to generate the channel feature vectors of feature maps. The detailed operation of feature embedding is shown in Fig. 5.

As shown in Fig. 5, the max-pooling and average-pooling operations are used to extract the global feature for each region. Moreover, the intra-region pixel adaptive weighting operation is also used to characterize the detail features. Finally, the extracted region features under the same channel are spliced to form the channel feature vector. Supposing the original feature map size is C×H×W, the dimension of the final embedded feature vector is $C \times d_k$, where $d_k = 3HW/S^2$.

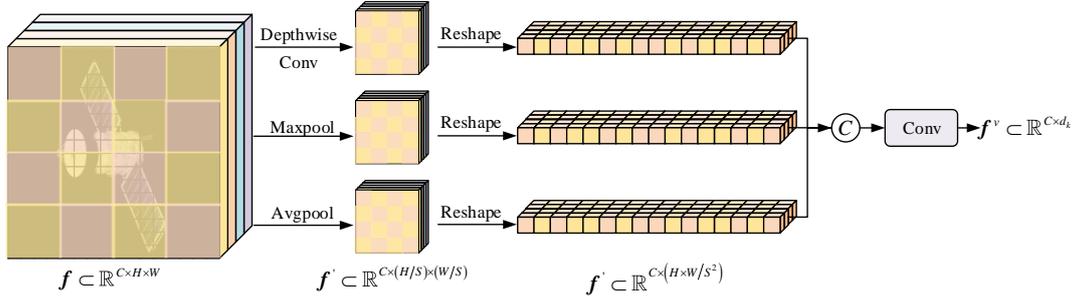

Fig. 5 The detailed operation of feature embedding. The max-pooling, average-pooing, and pixel adaptive weighting operations are utilized to characterize region features. Then the extracted region features under the same channel are spliced to form the channel feature vector.

After embedding the feature maps into feature vectors along the channel dimension, the multi-head co-attention mechanism is introduced to deduce the correlation between different features along the channel dimension. Assuming the feature maps of the gray image and depth map extracted by the convolutional neural network are $f_g \subset \mathbb{R}^{C \times H \times W}$ and $f_s \subset \mathbb{R}^{C \times H \times W}$, respectively, $f_g^v \subset \mathbb{R}^{C \times d_k}$ and $f_s^v \subset \mathbb{R}^{C \times d_k}$ are the embedded feature vector of $f_g$ and $f_s$, respectively. The query of $f_s^v$ and the key of $f_g^v$ can be calculated as

$$\begin{cases} Q_i = f_s^v W_i^Q, \ i \in [1, n] \\ K_i = f_g^v W_i^K, \ i \in [1, n] \end{cases} \tag{1}$$

where $W_i^Q \subset \mathbb{R}^{d_k \times (d_k/n)}$ and $W_i^K \subset \mathbb{R}^{d_k \times (d_k/n)}$ are the parameter matrices, $Q_i \subset \mathbb{R}^{C \times (d_k/n)}$ and $K_i \subset \mathbb{R}^{C \times (d_k/n)}$ are the query and the key of $f_s^v$ and $f_g^v$, respectively. $n$ is the number of attention heads, and $i$ is the index of attention heads. Then the cross-channel attention map $\omega_i \subset \mathbb{R}^{C \times C}$ between different features can be calculated as

$$\omega_i = \text{softmax}\left(\frac{Q_i K_i^\text{T}}{\sqrt{d_k}}\right) \tag{2}$$

The features of the gray image are then integrated into the depth map features according to the cross-channel attention, which can be expressed as

$$\tilde{f}_i = \text{reshape}\left(F_s^v + \omega_i F_g^v\right) \tag{3}$$

where $\text{reshape}(\cdot)$ denotes vector reshape operation, $F_s^v \subset \mathbb{R}^{C \times (H \times W)}$ and $F_g^v \subset \mathbb{R}^{C \times (H \times W)}$ are vectorized versions of $f_s$ and $f_g$, respectively. $\tilde{f}_i \subset \mathbb{R}^{C \times H \times W}$ represents the single-head fused feature, aggregating the grayscale and depth image information. Similar to Transformer [26], the multi-head attention mechanism is adopted to generate advanced fused features, which perform the single-head function in parallel. The inter-head fused features adaptive weighting is then adopted to enhance the representation ability of the fused features. The multi-head attention mechanism can be expressed as

$$\tilde{f}_s = \text{Conv}\left(\left[\tilde{f}_1; \tilde{f}_2; \ldots; \tilde{f}_n\right]\right) \tag{4}$$

where $n$ denotes the number of attention heads, and $[\cdot;\cdot]$ means concatenate operation. In this work, we set $n$ to 4.

The cross-channel attention layer can fully mine the correlation between heterologous feature maps along channel dimensions. However, it ignores the different information richness of the gray image features and the depth map features at different spatial locations. For instance, affected by lighting conditions, the shadow region in the gray image may fail to perceive spacecraft, while the LIDAR can still provide accurate ranging information in this region. On the other hand, in the regions with satisfactory lighting conditions, due to the sparseness of LIDAR ranging information, LIDAR loses the specific edge information. At this time, the gray image can provide clear edge features to guide spacecraft structure recovery. Therefore, the spatial attention layer is designed to retain more informative features at different spatial locations, which is complementary to the cross-channel attention module.

As shown in Fig. 4, the spatial attention layer takes the feature $\tilde{f}_s \subset \mathbb{R}^{C \times H \times W}$ output from the cross-channel attention layer and the gray image feature $f_g \subset \mathbb{R}^{C \times H \times W}$ as input and outputs the final fused feature $f_{out} \subset \mathbb{R}^{C \times H \times W}$ for subsequent depth prediction. Like CBAM [27], we first apply max-pooling and average-pooling operations along the channel axis on $\tilde{f}_s$ and $f_g$ and concatenate them to generate an efficient descriptor. Then a convolution layer and the sigmoid function are applied to generate the spatial attention map $W_{spatial} \subset \mathbb{R}^{1 \times H \times W}$, which encodes the richness of information at each location in the depth map. The final fused feature is the spatial-variant weighting of $\tilde{f}_s$ and $f_g$, which can be expressed as

$$f_{out} = W_{spatial} \odot \tilde{f}_s + (1 - W_{spatial}) \odot f_g \tag{5}$$

where $\odot$ denotes element-wise multiplication. During multiplication, the spatial attention map $W_{spatial}$ is broadcasted along the channel dimension.

C. Loss Functions

The loss functions of SDCNet consist of two parts: the foreground prediction loss for FSNet and the foreground depth completion loss for FDCNet. For the foreground prediction loss, the binary cross-entropy loss is adopted to supervise the foreground region segmentation, which can be expressed as

$$\mathcal{L}_{FSNet} = \frac{1}{N} \sum_i \left( -y_i \log(\hat{y}_i) - (1 - y_i) \log(1 - \hat{y}_i) \right) \tag{6}$$

where $y_i$ and $\hat{y}_i$ are the foreground ground truth label and predicted foreground probability, respectively. If the pixel belongs to the foreground region, its ground truth label is set to 1. Otherwise, it is set to 0.

The foreground depth completion loss aims to minimize the error between the network predicted depth and the ground truth depth in the predicted foreground region, which can be calculated as

$$\mathcal{L}_{FDCNet} = \frac{1}{N(\hat{A})} \sum_{x_i \in \hat{A}} |\hat{d}_i - d_i| \tag{7}$$

where $\hat{A}$ represents the set of pixels predicted as foreground pixels in FSNet. $N(\hat{A})$ denotes the number of elements in set $\hat{A}$, $\hat{d}_i$ and $d_i$ denote the predicted depth and ground truth depth for pixel $x_i$, respectively.

D. Evaluation Metrics

For the foreground segmentation subtask, we use Intersection over Union (IOU) and Intersection over Interest (IOI) to evaluate the performance of FSNet, which can be calculated as

$$IOU = \frac{N(A \cap \hat{A})}{N(A \cup \hat{A})} \tag{8}$$

$$IOI = \frac{N(A \cap \hat{A})}{N(A)} \tag{9}$$

where $\hat{A}$ denotes the set of pixels predicted as foreground pixels in FSNet, $A$ denotes the set of pixels belonging to the foreground in the ground truth label. $N(\cdot)$ is the function that counts the number of elements in the set. The IOI is mainly used to evaluate the retention rate of the ground truth foreground. The larger the IOI, the fewer foreground pixels are misclassified as background, and the more complete the retained foreground information is.

In the depth completion task, the mean absolute error (MAE) and the root mean squared error (RMSE) are often used to evaluate the performance of depth completion. However, since MAE and RMSE are the statistics of all pixel depth prediction errors, the results are easily unstable in the object-level depth completion task due to the change in the ratio of the foreground and background. Therefore, four new metrics to evaluate the quality of object-level depth completion results are proposed, including the mean absolute error of interest (MAEI), the mean absolute truncation error (MATE), the root mean square error of interest (RMSEI), and the root mean square truncation error (RMSTE), which can be calculated as

$$\text{MAEI} = \frac{1}{N(A \cap \hat{A})} \sum_{x_i \in A \cap \hat{A}} |\hat{d}_i - d_i| \tag{10}$$

$$\text{RMSEI} = \sqrt{\frac{1}{N(A \cap \hat{A})} \sum_{x_i \in A \cap \hat{A}} \|\hat{d}_i - d_i\|_2^2} \tag{11}$$

$$\text{MATE}(\alpha) = \frac{1}{N(\hat{A})} \sum_{x_i \in \hat{A}} \min(|\hat{d}_i - d_i|, \alpha) \tag{12}$$

$$\text{RMSTE}(\alpha) = \sqrt{\frac{1}{N(\hat{A})} \sum_{x_i \in \hat{A}} \min(\|\hat{d}_i - d_i\|_2^2, \alpha^2)} \tag{13}$$

where $\hat{A}$ and $A$ denote the set of pixels predicted as foreground pixels and the set of pixels belonging to the foreground in the ground truth, respectively. $\hat{d}_i$ and $d_i$ denote the predicted depth and ground truth depth for pixel $x_i$. $N(\cdot)$ is the function that counts the number of elements in the set, and the parameter $\alpha$ is the truncation threshold.

MAEI, RMSEI, MATE, and RMSTE essentially calculate MAE and RMSE in different image areas. MAEI and RMSEI calculate MAE and RMSE in the intersection area of the predicted foreground and the ground truth foreground, which can more intuitively reflect the depth completion accuracy of the spacecraft itself. On the other hand, MATE and RMSTE calculate MAE and RMSE in the predicted foreground area. Considering the depth error of the region misclassified as the foreground is very large and depends on the object's distance, threshold truncation is performed to prevent it from causing excessive influence on MAE and RMSE. In our work, we set the truncation threshold $\alpha$ to 10m.

## IV. SATELLITE DEPTH COMPLETION DATASET CONSTRUCTION

Training a deep network for spacecraft depth completion requires an extensive collection of satellite images and LIDAR data. To date, there is no public dataset for spacecraft depth completion tasks. This paper constructs a large-scale satellite depth completion dataset for training and testing spacecraft depth completion algorithms.

We performed the imaging simulation of spacecraft using blender software based on 126 satellite CAD models. The specific camera and LIDAR parameters are shown in Table 1 and Table 2. Specifically, the LIDAR parameters are derived from the JAGUAR product of Innovusion company. During the simulation, we randomly set the satellite solar plane size to 3-8m and the body size to 1-3m. The satellite's distance to the observation platform is set in the range of 50-250m, and the satellite's attitude relative to the observation platform is random. The maximum angle between the illumination and observation angles is set to 70°.

Table 1 The specifications of the optical camera

| Parameter | Value |
|---|---|
| Focal length/mm | 50 |
| Field of view/° | 7.38×7.38 |
| Image resolution/pixel | 1024×1024 |
| Sensor size/mm | 6.449×6.449 |

Table 2 The specifications of the LIDAR

| Parameter | Value |
|---|---|
| Maximum range/m | 280 |
| Range error/cm | <3 |
| Vertical angular resolution/° | 0.13 |
| Horizontal angular resolution/° | 0.09 |

We simulated 64 sets of data under different observation conditions for each satellite model, resulting in 8064 sets of gray images and LIDAR depth maps. Fig. 6 shows some simulation results of different satellite models. To ensure the generalization performance of the algorithm to different satellite models, we divided the simulation data of 126 satellite models into three subsets: training set (simulation results of 99 models), validation set (simulation results of 9 models), and test set (simulation results of 18 models), resulting in 6336, 576 and 1152 sets of data for training, validation, and testing. In this way, we can ensure that the data used for validation and testing is invisible to the depth completion network, putting forward high requirements for the applicability of the algorithm to unknown satellite models.

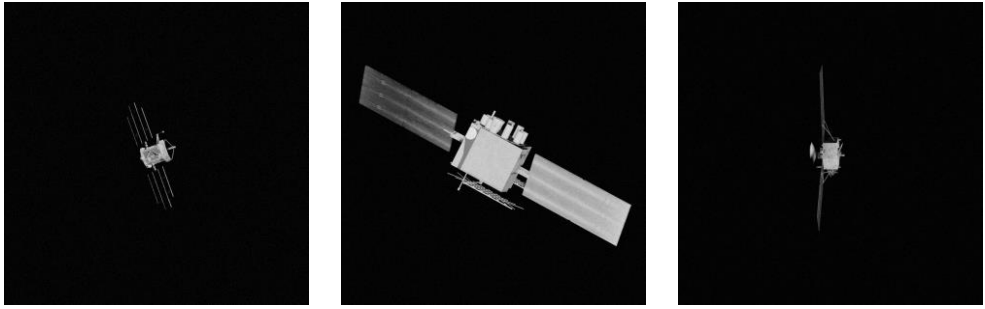

(a) The optical camera imaging simulation results

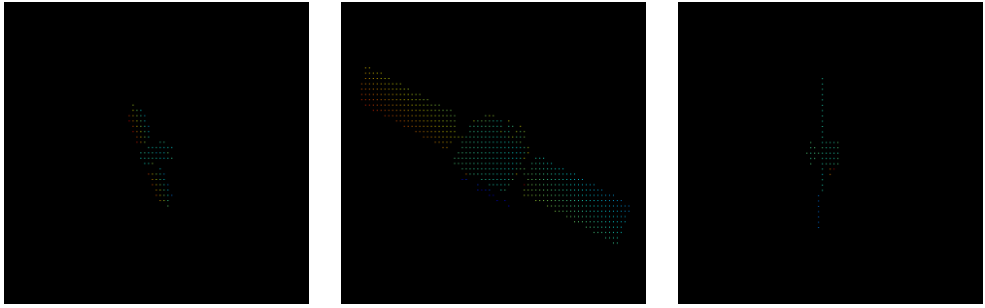

(b) The LIDAR sparse ranging simulation results

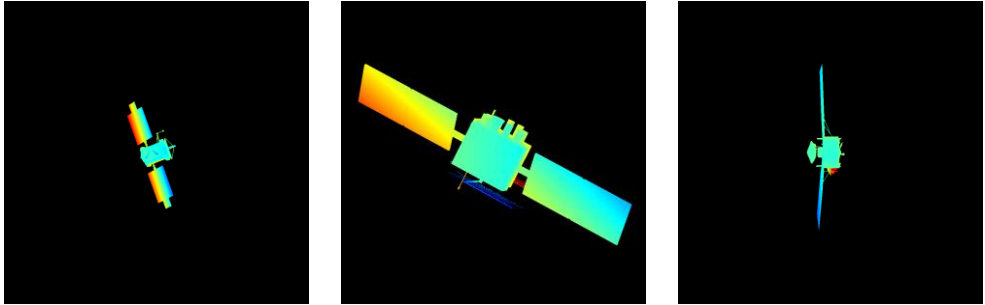

(c) The ground truth dense depth map

Fig. 6 Examples of simulation results of different satellite models. From top to bottom: the gray image, the sparse depth map which is dilated for better visualization, and the ground truth dense depth map.

## V. EXPERIMENTS

A. Experiment Setup

The proposed SDCNet is implemented in python using the Paddle library and trained on an Nvidia Tesla V100 GPU. During training, we first train the FSNet using the standard binary cross-entropy loss, then freeze the weights in FSNet and train the whole network. We randomly crop or resize images to 512×512 to train effectively with limited computational resources. We train the FSNet and the whole model using the Adam [28] optimizer for 30 and 50 epochs, with an initial learning rate of 0.001 and a weight decay of 0.001. In addition, data augmentation techniques, including random flip and image jitter, are adopted.

B. Experiment Results

We compare our method with state-of-the-art depth-completion methods on the constructed satellite depth-completion dataset, including Sparse-to-dense [7], CSPN [15], GuideNet [20], FCFRNet [21], RigNet [22], PENet [13], and DySPN [18]. All the methods are trained on the same training set and evaluated on the same test set. It should be noted that since the spacecraft LIDAR depth maps are more sparse compared to ground scenes, most of the existing depth completion methods fail to predict reasonable dense depth results under the input of the original sparse depth map. To this end, on the basis of the original method, we additionally adopt the morphological preprocessing operation on the sparse depth map to generate the pseudo-dense depth map and feed it into the depth completion network.

Table 3 The quantitative results of different depth completion methods. All the results tested on existing methods are obtained by adding the morphological preprocessing operation based on the original method to ensure reasonable completion results. Bold values represent the optimal values of different evaluation criteria among all methods.

| Methods | MAEI/m | MATE/m | RMSEI/m | RMSTE/m |
|---|---|---|---|---|
| Sparse-to-dense | 3.193 | 5.442 | 4.227 | 6.758 |
| CSPN | 5.033 | 8.455 | 6.176 | 9.001 |
| GuideNet | 0.620 | 1.183 | 1.271 | 2.614 |
| FCFRNet | 1.180 | 1.826 | 1.893 | 3.133 |
| RigNet | 0.448 | 1.748 | 1.092 | 3.670 |
| PENet | 0.403 | 1.005 | 0.885 | 2.568 |
| DySPN | 0.389 | 0.934 | 0.940 | 2.426 |
| SDCNet | **0.250** | **0.759** | **0.627** | **2.229** |

Table 3 lists the quantitative results of different methods. As shown in Table 3, compared to the state-of-the-art depth-completion methods, the SDCNet proposed in this paper outperforms all other methods in all indicators. Fig. 7 shows some qualitative depth completion examples of SDCNet. The first and second rows in Fig. 7 are the gray images and the sparse depth maps, respectively. They are the input of the depth completion network. The third row in Fig. 7 shows the foreground segmentation results predicted by FSNet. It can be seen that thanks to the complementary information of gray images and depth maps, the designed lightweight foreground segmentation can accurately segment the foreground region even under unfavorable illumination conditions. The fourth and fifth rows in Fig. 7 are the ground truth dense depth map and the predicted dense depth map, respectively. The last row in Fig. 7 also shows the point cloud generated from the predicted depth map for better visualization. It can be seen that the generated point cloud can accurately recover the three-dimensional structure of the spacecraft, which verifies the effectiveness of the proposed method in this paper.

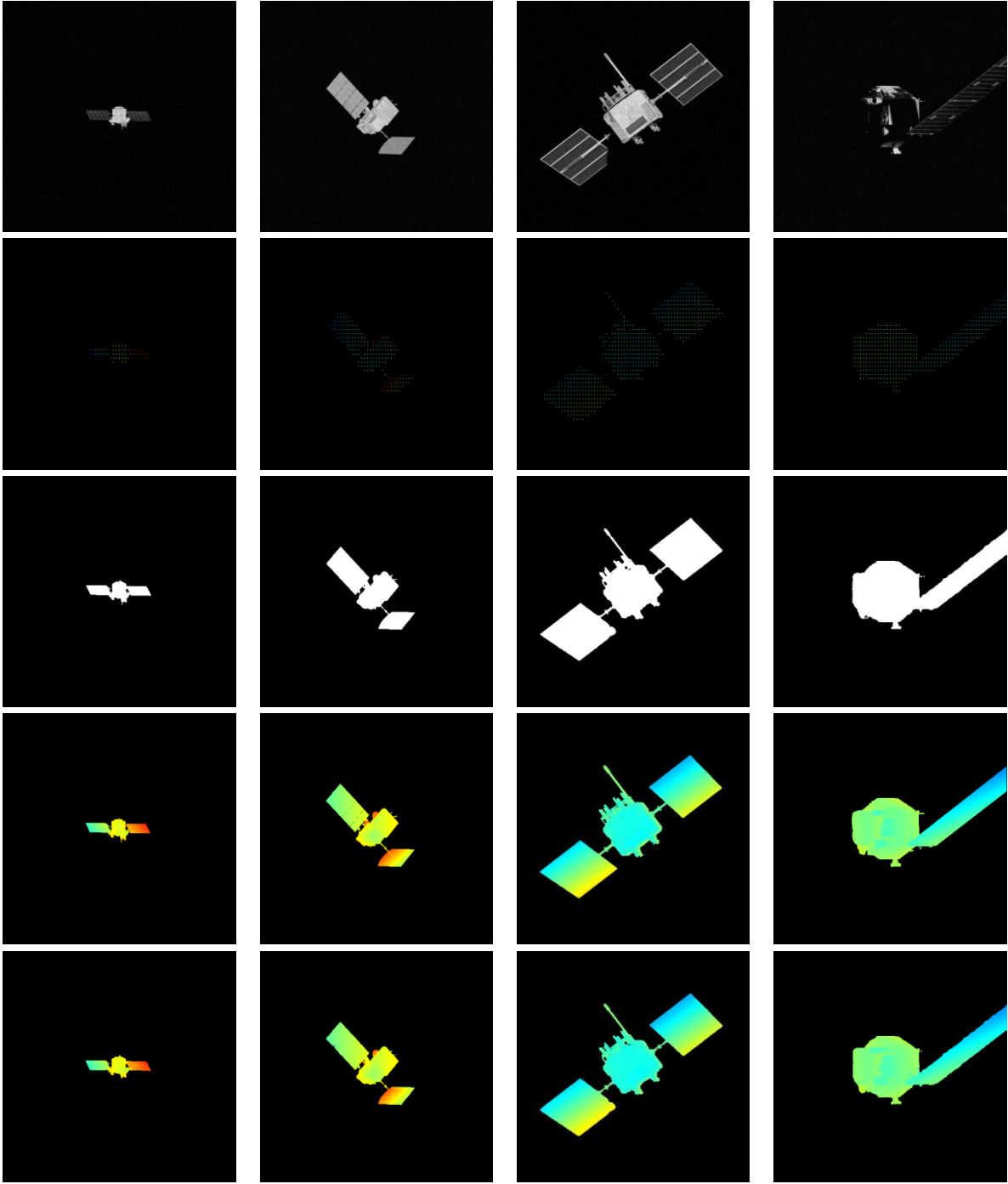

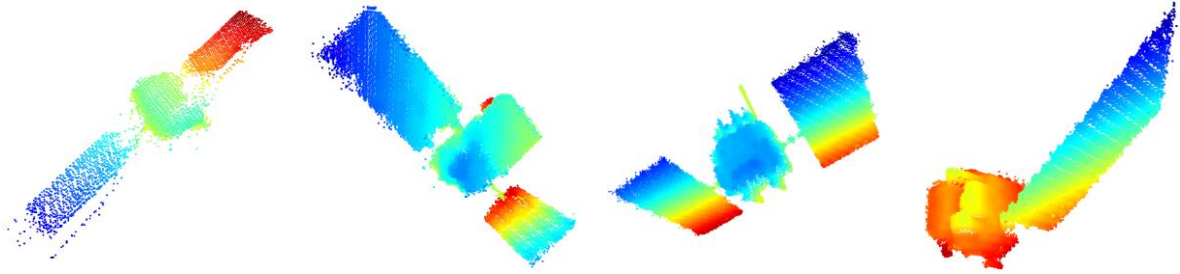

Fig. 7 Some qualitative depth completion examples. From top to bottom: the gray image, the sparse depth map which is dilated for better visualization, the foreground segmentation result predicted by FSNet, the ground truth dense depth map, the predicted depth map, the point cloud converted from predicted depth map (statistical outlier removal operation is adopted for better point cloud quality).

C. Ablation Studies

In this section, we first conduct additional experiments to explore how different input choices affect the performance of the FSNet. Moreover, ablation studies are also conducted to verify the effectiveness of each component proposed in our method, including the foreground prediction network (FSNet), the cross-channel attention layer (CCA), and the spatial attention layer (SA).

1. Effects of different inputs to FSNet

In order to analyze the influence of different inputs on the performance of the foreground prediction network, we train the FSNet with different inputs, and the quantitative experimental results are listed in Table 4.

Table 4 The quantitative results of FSNet under different input choices. Bold values represent the optimal values of different evaluation criteria among all versions.

| gray image | sparse depth map | IOU/% | IOI/% |
|---|---|---|---|
| √ |   | 86.73 | 90.80 |
|   | √ | 86.89 | 93.22 |
| √ | √ | **95.30** | **97.44** |

As shown in Table 4, when the FSNet solely takes the gray image as input, this version of FSNet performs worst, and the IOU and IOI of the prediction results are only 86.73% and 90.8%. That is because the on-orbit lighting condition is complex, and the regions in the shadow are generally invisible to the gray images. When the FSNet solely takes the sparse depth map as input, the IOU and IOI increase to 86.89% and 93.22%, respectively. That is because the

LIDAR is not sensitive to illumination, and the LIDAR can still provide ranging information even if the illumination conditions are poor. However, the LIDAR can not provide the edge structure information due to its sparse ranging results, resulting in a 2.42% increase in IOI while almost no improvement in IOU. The final version takes the gray image and the sparse depth map jointly into the FSNet, and prediction results are significantly improved, with the IOU and IOI reaching 95.3% and 97.44%, respectively. On the one hand, the depth map can provide distance information for the areas where the gray image features are annihilated. On the other hand, the gray image can provide clear edge structure information in satisfactory lighting conditions. In this way, FSNet can take full advantage of the complementary information of different sensor data and accurately classify whether a pixel belongs to the foreground under different lighting conditions.

2. Effectiveness of different component

We choose the state-of-the-art depth completion network GuideNet [20] as the baseline, which has the same structure as FDCNet except employs the guided convolution module to fuse the features extracted from the optical image and the depth map. On this basis, the FSNet is first introduced to enable the network to focus on recovering the foreground region depth (the second version in Table 5). It can be seen that the introduction of the FSNet significantly improves depth completion accuracy. This phenomenon reflects that the background pixels will interfere with the depth recovery of the object itself and degrade the network performance. At the same time, it verifies the effectiveness of decomposing the object-level depth completion task into the foreground segmentation subtask and foreground depth completion subtask. Further replacing the guided convolution module with the proposed cross-channel attention layer (the third version in Table 5), the depth prediction error MAEI and MATE are further reduced by 3.7cm and 5cm, respectively. Similarly, replacing the guided convolution module with the spatial attention layer can also decrease the depth completion error (the fourth version in Table 5). The final version (the fifth version in Table 5), which successively uses the cross-channel and spatial attention layers to aggregate features of different inputs, achieves the best performance, verifying the effectiveness of the multi-source feature fusion module.

Table 5 The quantitative results of SDCNet with different components. Bold values represent the optimal values of different evaluation criteria among all versions.

| baseline | FSNet | CCA | SA | MAEI/m | MATE/m | RMSEI/m | RMSTE/m |
|---|---|---|---|---|---|---|---|
| √ | | | | 0.620 | 1.183 | 1.271 | 2.614 |
| √ | √ | | | 0.338 | 0.859 | 0.751 | 2.297 |
| √ | √ | √ | | 0.301 | 0.809 | 0.747 | 2.269 |
| √ | √ | | √ | 0.295 | 0.811 | 0.765 | 2.288 |
| √ | √ | √ | √ | **0.250** | **0.759** | **0.627** | **2.229** |

D. Application for Pose Estimation

To evaluate the predicted depth quality and explore the feasibility of utilizing the predicted dense depth map in downstream vision tasks, we conduct the spacecraft pose estimation experiment based on the results of SDCNet. Specifically, we simulated a sequence of data containing 36 frames for each satellite in the test set, and the Euler angle and position along each axis are randomly increased within [5°,10°] and [1m,2m] for each adjacent frame. The simulated data is then fed into SDCNet to obtain the predicted dense map, which is used as the input for the pose estimation experiments.

We choose the state-of-the-art spacecraft pose estimation method PANet[2] as our pose estimation method in this experiment. Since PANet requires the point clouds of spacecraft as input, we convert the predicted depth map into the point cloud according to the camera parameters, and the statistical outlier removal operation is adopted to improve point cloud quality. Fig. 8 shows several pose estimation visualization examples on the testing point cloud. It can be seen that the transformed source point cloud (the point clouds in red) aligns well with the target point cloud (the point clouds in blue), which implies the pose estimation results are with high accuracy. According to statistics, the average three-axis rotation and translation errors are 0.85° and 1.47m, respectively, which verified that the predicted dense depth could be applied in downstream vision tasks.

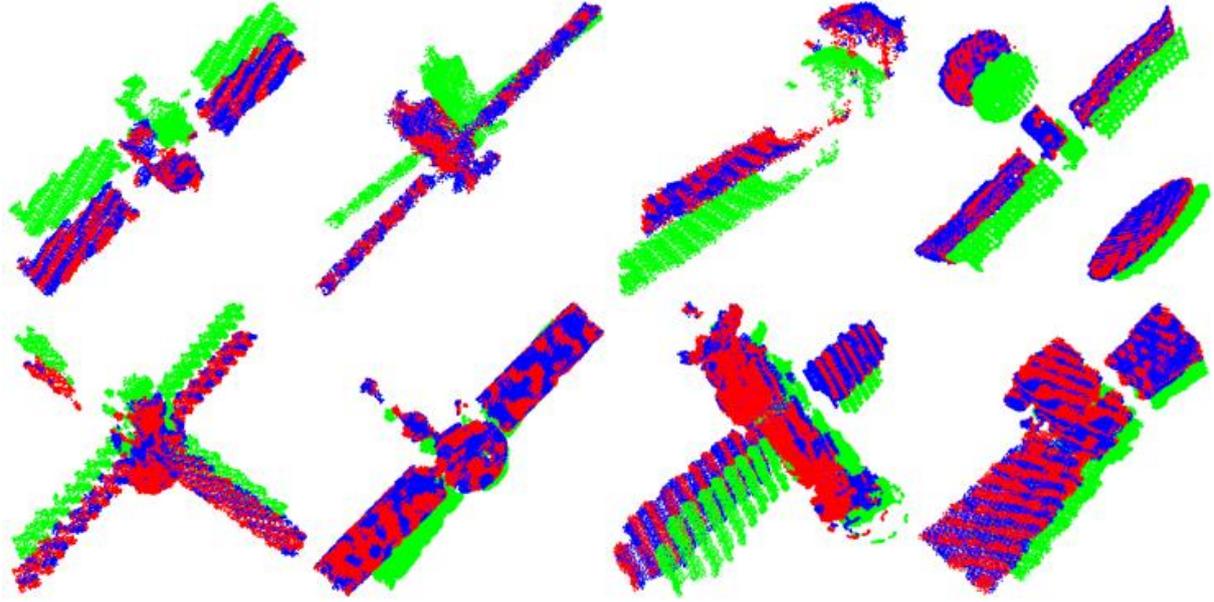

Fig. 8 Some qualitative examples of spacecraft pose estimation. The green, blue, and red point clouds denote the source, target, and transformed source point cloud, respectively.

## VI. CONCLUSIONS

Aiming at the limited work distance of the existing stereo vision system and active time-of-flight (TOF) camera, this paper proposes to sense the three-dimensional structure of spacecraft at a long distance (maximum to 250m) using LIDAR and a monocular camera. To this end, a novel Spacecraft Depth Completion Network (SDCNet) is proposed to recover the dense depth map using a gray image and sparse depth map. Considering that the celestial background inevitably interferes with the spacecraft depth recovery, the object-level spacecraft depth completion task is decomposed into the foreground segmentation subtask and the foreground depth completion subtask. Specifically, a lightweight foreground segmentation subnet (FSNet) is designed for foreground region segmentation first, and the pixel's depth in the segmented region is regressed using the foreground depth completion subnet (SDCNet). Moreover, we design the attention-based feature fusion module to deduce the correlation between different features along the channel and the spatial dimension sequentially, integrating the geometric features and context the gray image provides into the depth map feature. Four new metrics for the object-level depth completion task are also proposed to evaluate depth completion results, including MAEI, MATE, RMSEI, and RMSTE. Besides, we construct a large-scale satellite depth completion dataset based on 126 satellite CAD models, containing 6336, 576, and 1152 sets of data for training, validation, and testing the spacecraft depth completion algorithms. The construction of the satellite depth completion dataset solves the lack of satellite data for training and testing depth completion methods. Empirical experiments on the dataset demonstrate that our method achieves state-of-the-art depth completion performance, which achieves 0.25m mean absolute error of interest and 0.759m mean absolute truncation error. Finally, the spacecraft pose estimation experiment is also conducted based on the depth completion results, which achieves 0.85° rotation error and 1.47m translate error, verifying that the predicted depth map could meet the needs of downstream vision tasks. The proposed spacecraft depth completion method has the potential to be integrated into space on-orbit service systems, which can perceive the fine three-dimensional structure of spacecraft at a long distance using LIDAR and optical camera.

## VII REFERENCES

[1] J. Ventura J et al., "Pose tracking of a noncooperative spacecraft during docking maneuvers using a time-of-flight sensor," in AIAA Guidance, Navigation, and Control Conference, California, USA, 2016, pp. 0875, doi: 10.2514/6.2016-0875.

[2] X. Liu et al., "Position Awareness Network for Non-Cooperative Spacecraft Pose Estimation Based on Point Cloud,". IEEE Transactions on Aerospace and Electronic Systems, early access, 2022, doi: 10.1109/TAES.2022.3182307.

[3] Q. Wei et al., "Robust spacecraft component detection in point clouds,". Sensors, vol. 18, no. 4, pp. 933, Mar. 2018, doi: 10.3390/s18040933.

[4] R. Volpe et al., "Reconstruction of the Shape of a Tumbling Target from a Chaser in Close Orbit," in IEEE Aerospace Conference, MT, USA, 2020, pp. 1-11, doi: 10.1109/AERO47225.2020.9172529.

[5] W. Xu et al., "A pose measurement method of a non-cooperative GEO spacecraft based on stereo vision," in International Conference on Control Automation Robotics & Vision, Guangzhou, China, 2012, pp. 966-971, doi: 10.1109/ICARCV.2012.6485288.

[6] H. G. Martínez et al., "Pose estimation and tracking of non-cooperative rocket bodies using time-of-flight cameras," Acta Astronautica, vol. 139, pp. 165-175, Oct. 2017, doi: 10.1016/j.actaastro.2017.07.002.

[7] F. Ma et al., "Sparse-to-dense: Depth prediction from sparse depth samples and a single image," in IEEE international conference on robotics and automation, Brisbane, Australia, 2018, pp. 4796-4803, doi: 10.1109/ICRA.2018.8460184.

[8] S. Imran et al., "Depth coefficients for depth completion," in IEEE/CVF Conference on Computer Vision and Pattern Recognition, Long Beach, USA, 2019, pp. 12438-12447, doi: 10.1109/CVPR.2019.01273.

[9] F. Ma et al., "Self-supervised sparse-to-dense: Self-supervised depth completion from lidar and monocular camera," in IEEE international conference on robotics and automation, Montreal, Canada, 2019, pp. 3288-3295, doi: 10.1109/ICRA.2019.8793637.

[10] B. U. Lee et al., "Depth completion using plane-residual representation," in IEEE IEEE/CVF Conference on Computer Vision and Pattern Recognition, Nashville, USA, 2021, pp. 13911-13920, doi: 10.1109/CVPR46437.2021.01370.

[11] P. Hambarde P et al., "S2DNet: Depth estimation from single image and sparse samples," IEEE Transactions on Computational Imaging, vol. 6, pp. 806-817, Mar. 2020, doi: 10.1109/TCI.2020.2981761.

[12] L. Liu et al., "Learning steering kernels for guided depth completion," IEEE Transactions on Image Processing, vol. 30, pp. 2850-2861, Feb. 2021, doi: 10.1109/TIP.2021.3055629.

[13] M. Hu et al., "Penet: Towards precise and efficient image guided depth completion," in IEEE international conference on robotics and automation, Xi'an, China, 2021, pp. 13656-

13662, doi: 10.1109/ICRA48506.2021.9561035.

[14] S. Liu et al., "Learning affinity via spatial propagation networks," in Proccedings of Neural Information Processing Systems, Long Beach, USA, 2017, pp. 1519-1529, doi: 10.48550/arXiv.1710.01020.

[15] X. Cheng et al., "Learning depth with convolutional spatial propagation network,". IEEE Transactions on Pattern Analysis and Machine Intelligence, vol. 42, no. 10, pp. 2361-2379, Oct. 2019, doi: 10.1109/TPAMI.2019.2947374.

[16] X. Cheng et al., "Cspn++: Learning context and resource aware convolutional spatial propagation networks for depth completion," in Proceedings of the AAAI Conference on Artificial Intelligence, New York, USA, 2020, pp. 10615-10622, doi: 10.1609/aaai.v34i07.6635.

[17] J. Park et al., "Non-local spatial propagation network for depth completion," in Proceedings of the European conference on computer vision, 22020, pp. 120-136, doi: 10.1007/978-3-030-58601-0_8.

[18] Y. Lin et al., "Dynamic spatial propagation network for depth completion,". 2022, arXiv:2202.09769.

[19] K. He et al., "Guided image filtering," IEEE Transactions on Pattern Analysis and Machine Intelligence, vol. 35, no. 6, pp. 1397-1409, Oct. 2012, doi: 10.1109/TPAMI.2012.213.

[20] J. Tang et al., "Learning guided convolutional network for depth completion," IEEE Transactions on Image Processing, vol. 30, pp. 1116-1129, Dec. 2020, doi: 10.1109/TIP.2020.3040528.

[21] L. Liu et al., "Fcfr-net: Feature fusion based coarse-to-fine residual learning for depth completion," in Proceedings of the AAAI Conference on Artificial Intelligence, 2021, pp. 2136-2144, doi: 10.48550/arXiv.2012.08270.

[22] Z. Yan et al., "RigNet: Repetitive image guided network for depth completion," 2021,

arXiv:2107.13802.

[23] Y. Chen et al., "Learning joint 2d-3d representations for depth completion," in Proceedings of the IEEE/CVF International Conference on Computer Vision, Seoul, Korea, 2019, pp. 10023-10032, doi: 10.1109/ICCV.2019.01012.

[24] J. Uhrig et al., "Sparsity invariant cnns," in Proceedings of the international conference on 3D Vision, Qingdao, China, 2017, pp. 11-20, doi: 10.1109/3DV.2017.00012.

[25] Z. Liu et al., "Swin transformer: Hierarchical vision transformer using shifted windows," in Proceedings of the IEEE/CVF International Conference on Computer Vision, QC, Canada, 2021, pp. 9992-10002, doi: 10.1109/ICCV48922.2021.00986.

[26] A. Vaswani et al., "Attention is all you need,". in Proccedings of Neural Information Processing Systems, Long Beach, USA, 2017, pp. 6000-6010. doi: arXiv:1706.03762.

[27] S. Woo et al., "Cbam: Convolutional block attention module," in Proceedings of the European conference on computer vision, Munich, Germany, 2018, pp. 3-19, doi: 10.48550/arXiv.1807.06521.

[28] D. P. Kingma et al., "Adam: A method for stochastic optimization," 2014, arXiv:1412.6980.